\pdfoutput=1

\documentclass[11pt]{article}

\usepackage[preprint]{acl}

\usepackage{times}
\usepackage{latexsym}

\usepackage[T1]{fontenc}

\usepackage[utf8]{inputenc}

\usepackage{microtype}

\usepackage{inconsolata}
\usepackage{soul}

\usepackage{graphicx}
\usepackage{stfloats}
\usepackage{subfigure}
\usepackage{booktabs}
\usepackage{multirow}

\title{Talk With Human-like Agents: Empathetic Dialogue Through Perceptible Acoustic Reception and Reaction}

\newcommand\CoAuthorMark{\footnotemark[\arabic{footnote}]}

\author{Haoqiu Yan\thanks{Equal contribution.}$^{1,3}$, Yongxin Zhu\CoAuthorMark$^{2,3}$, Kai Zheng$^{1,3}$, \AND Bing Liu$^{4}$, Haoyu Cao$^{4}$, Deqiang Jiang$^{4}$, Linli Xu\thanks{Corresponding author.}$^{1,3} $ \\
$^{1}$School of Computer Science and Technology, University of Science and Technology of China \\
$^{2}$School of Data Science, University of Science and Technology of China \\
$^{3}$State Key Laboratory of Cognitive Intelligence, 
$^{4}$Tencent Youtu Lab \\
\texttt{\{yanhq,zyx2016,dthdzk\}@mail.ustc.edu.cn} \\
\texttt{\{billbliu,rechycao,dqiangjiang\}@tencent.com  }   
\texttt{linlixu@ustc.edu.cn}
}

\begin{document}

\maketitle
\begin{abstract}


Large Language Model (LLM)-enhanced agents become increasingly prevalent in Human-AI communication, offering vast potential from entertainment to professional domains. However, current multi-modal dialogue systems overlook the acoustic information present in speech, which is crucial for understanding human communication nuances. This oversight can lead to misinterpretations of speakers' intentions, resulting in inconsistent or even contradictory responses within dialogues. To bridge this gap, in this paper, we propose PerceptiveAgent, an empathetic multi-modal dialogue system designed to discern deeper or more subtle meanings beyond the literal interpretations of words through the integration of speech modality perception. Employing LLMs as a cognitive core, PerceptiveAgent perceives acoustic information from input speech and generates empathetic responses based on speaking styles described in natural language.  Experimental results indicate that PerceptiveAgent excels in contextual understanding by accurately discerning the speakers' true intentions in scenarios where the linguistic meaning is either contrary to or inconsistent with the speaker's true feelings, producing more nuanced and expressive spoken dialogues. Code is publicly available at: \url{https://github.com/Haoqiu-Yan/PerceptiveAgent}.

\end{abstract}

\section{Introduction}

Artificial Intelligence (AI) agents \citep{ai-intro1,ai-intro2} are entities designed to replicate human-like intelligence and functionalities, serving as the essential building blocks of AI systems. An ideal agent should be capable of perceiving its environment with sensors, making informed decisions, and then taking actions in response to users or scenarios. Recently, Large Language Models (LLMs) \citep{llm-intro1,llm-intro2,llm-intro3} have exhibited remarkable capabilities in diverse tasks, offering opportunities for building general AI agents that engage in human-like interactions, such as virtual assistants and intelligent robots. However, current text-only dialogue systems \citep{chatgpt,llama2} fall short in bridging the gap between experimental and realistic scenarios, where humans perceive and understand the world through diverse multi-modal information. Thus, the integration of acoustic information into dialogues has the potential to foster the development of more human-like agents, thereby enhancing the empathetic experience they offer.

\begin{figure}[tp]
    \centering
    \includegraphics[width=1.1\linewidth]{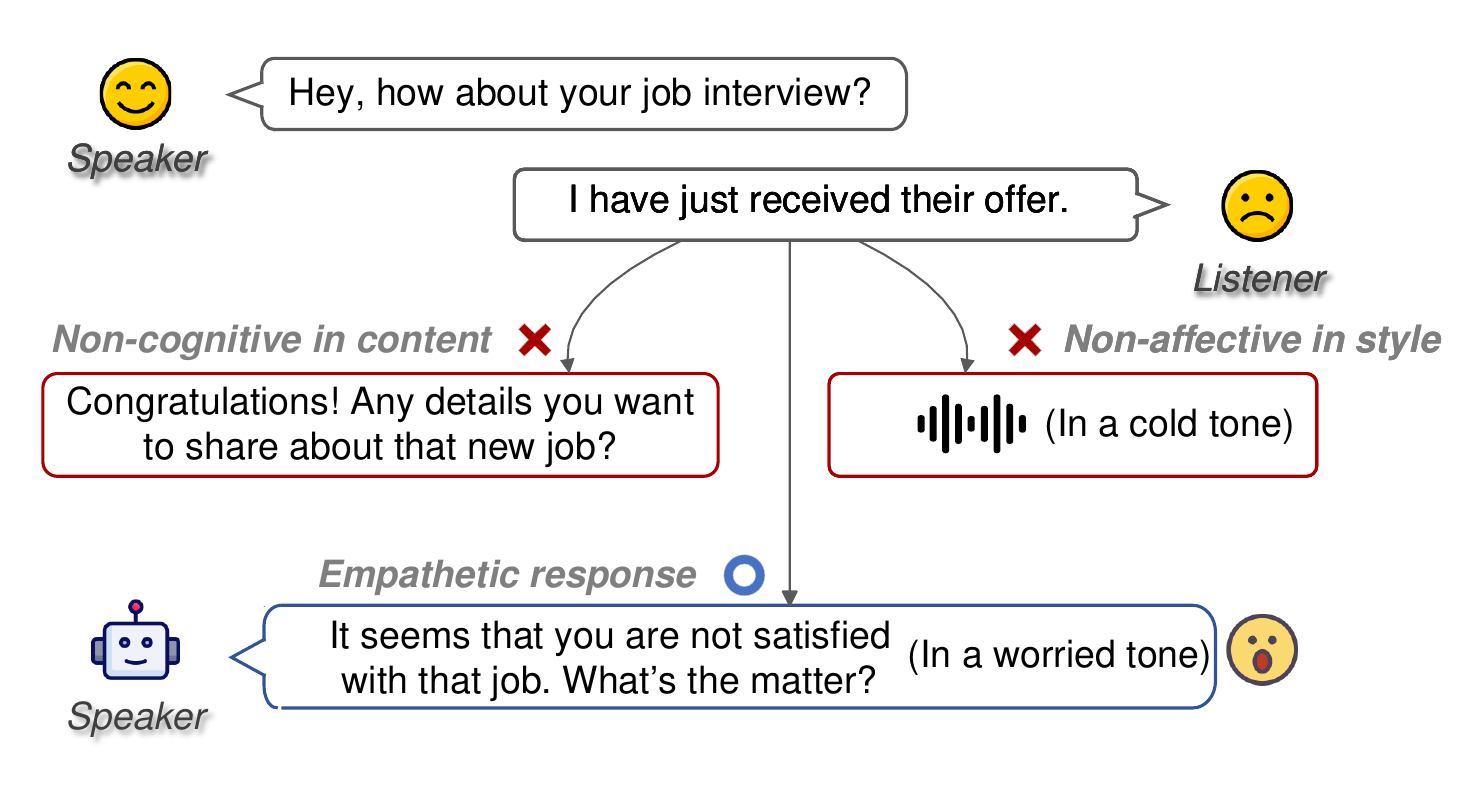}
    \caption{Examples illustrating the definition of empathy within dialogues.}
    \label{fig:story}
\end{figure}

Empathetic responses involve two essential aspects: cognitive and affective empathy \citep{empathetic-cuff2016empathy,empathetic-kim2021perspective,empathetic-reis2011familiarity,empathetic-smith2006cognitive}, which reflect an understanding of the human-talker's thoughts and feelings respectively. Specifically, cognitive empathy involves understanding the human-talker's thoughts, perspectives, and described events, enabling the agent to provide responses relevant to the dialogue topic~\citep{sabour2022cem}. Conversely, affective empathy entails responding based on observed emotional expressions in the dialogue history, contributing to the naturalness of synthesized speech \citep{controllable,voiceagents,acoustic}. While recent works \citep{chatgpt-edss,dgslm,written} leverage LLM's  strong capabilities of contextual understanding and content generation to synthesize empathetic speeches, there remains a discrepancy between cognitive and affective empathy. This arises because cognitive content is preassigned before affective speech is deduced from latent representations of multi-modal dialogue history.

Recently, advancements in multi-modal content perception and generation have been achieved by various methods ~\citep{zhang2023speechgpt,huang2023audiogpt,chen2023xllm,wu2023nextgpt}, where audio is represented as either recognized text with an automatic speech recognition model or discrete features with a speech encoder. However, while linguistic information in speech is predominantly captured by both discrete acoustic units and textual representations, acoustic features tend to be disregarded. This oversight can lead to misinterpretations of the speaker's intentions, resulting in discrepant or even contradictory responses within the dialogue history. As illustrated in Figure \ref{fig:story}, the left scenario fails to consider the perspective of the listener while the right one barely understands or empathizes with the speaker's feelings.


In this paper, we propose \textbf{PerceptiveAgent}, an empathetic multi-modal dialogue system that can discern deeper or more subtle meanings beyond the literal interpretations of words, based on speaking styles described in natural language. Specifically, PerceptiveAgent first comprehends the speaker's intentions accurately by a perceptive captioner model that captures acoustic features from each speech within dialogues. Subsequently, an LLM module acts as the cognitive core, producing the relevant response content with a caption describing how to articulate the response. A Multi-Speaker and Multi-Attribute Synthesizer (MSMA-Synthesizer) is then developed to synthesize nuanced and expressive speech. 

Our contributions include the following:
\begin{itemize}
\item[$\bullet$] We pioneer the construction of a speech captioner model to perceive and express acoustic information through natural language. 
\item[$\bullet$] We develop an empathetic multi-modal dialogue system capable of identifying the speaker's true intentions through audio modality perception and generating empathetic speech.
\item[$\bullet$] Experiments demonstrate that PerceptiveAgent can accurately discern the true intentions in scenarios where the literal interpretations of words are either contrary to or inconsistent with the speaker's true feelings.
\end{itemize}

\section{Related Work}

\subsection{Multi-modal Dialogue Systems} Recent advances in multi-modal dialogue systems have primarily focused on transforming speech into discrete latent representation. For instance, \citet{zhang2023speechgpt,chen2023xllm,wu2023nextgpt} utilize  speech encoders to perceive speech and then synthesize responses according to discrete acoustic units derived from LLMs, showing intrinsic cross-modal conversational abilities. Besides, works including~\citep{dgslm,written} 
autonomously generate two-channel spoken dialogues, simulating realistic interactions between agents, including vocal interactions, laughter, and turn-taking. However, while discrete acoustic units capture linguistic information effectively, prosodic features are mostly ignored. To address this limitation and preserve prosodic information as much as possible, we develop a multi-modal dialog system that perceives prosody through speech captioning and responds empathetically using an LLM and a speech synthesizer. 

\subsection{Cross-Modal Text Generation} Cross-modal text generation involves generating text conditioned on other modalities such as audio and vision~\citep{li2022blip,chao2024cvpr,zhang2024visual}, where the key challenge is to align multi-modal features with the text latent space. Recent approaches ~\citep{zhu2023minigpt,chen2023xllm} address this challenge by aligning off-the-shelf pre-trained LLMs with learnable visual encoders \citep{li2023blip,zhao2023bubogpt}, transforming multi-modal representations as learnable query embeddings while keeping both pre-trained LLMs and visual encoders frozen. Similarly, for audio caption tasks, audio embeddings are mapped to a sequence of prefix vectors and then taken as the context input for caption generation \citep{beyond57,beyond58,xu24secap}. However, to the best of our knowledge, we are the first to construct a speech captioner capable of perceiving acoustic information in dialogues.

\subsection{Expressive Text-to-Speech Synthesis} Given a transcript, text-to-speech (TTS) models achieve voice variability by conditioning on a zero-shot speech prompt or a text prompt of the desired style. For instance, zero-shot TTS systems reproduce the speaker characteristics and acoustic environments of a speech prompt through in-context learning \citep{pt2-1,pt2-2,pt2-3,pt2-4} . However, these systems lack independent control over speaking styles, including prosody, emotion, and acoustic environment. To address this, text prompts have been introduced for more natural and general speech synthesis. Approaches like~\citep{guo2023prompttts,leng2023prompttts,shimizu2023prompttts++,ji2023textrolspeech} 
express speaking styles in natural language, while methods such as \citep{polyak2021speech,nguyen2023expresso} 
utilize explicit labels to generate diverse speech 
that matches the prompt. We follow the latter direction and construct a speech synthesis model with multiple speaking style labels.

\section{Methods}

\begin{figure*}[ht!]
    \vspace{-1cm}
    \centering
    \includegraphics[width=1\linewidth]{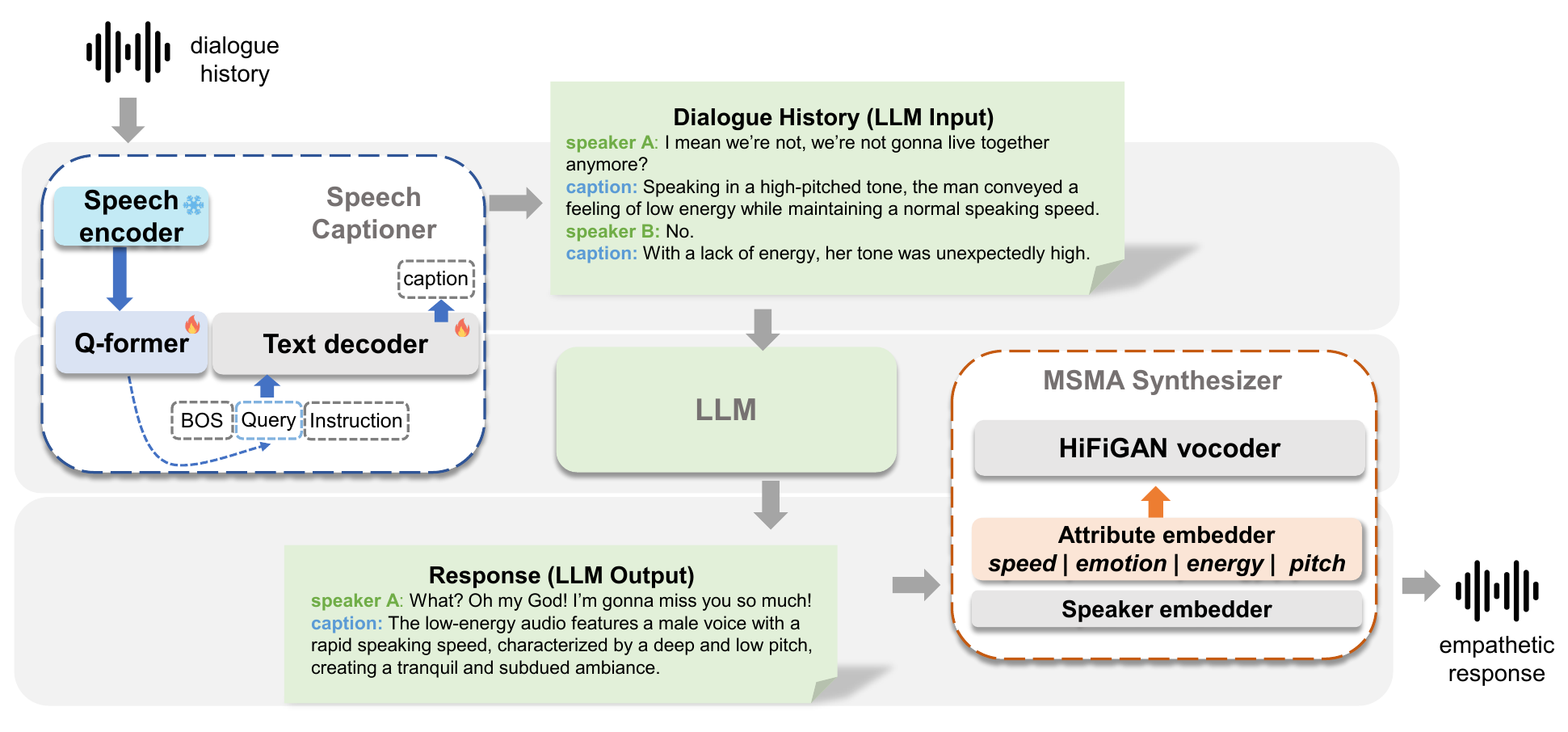}
    \caption{The overall architecture of 
    PerceptiveAgent. Three components are interconnected: the speech captioner, the LLM and the MSMA-Synthesizer. The speech captioner serves as a multi-modal sensory system, perceiving acoustic information from the dialogue history, which is crucial for discerning the speakers' intentions. The LLM acts as the cognitive core, responsible for comprehending the speakers' thoughts and emotions. Conditioned on the response contents and multiple attributes provided by the LLM, the MSMA-Synthesizer generates expressive speech outputs.}
    \label{fig:model}
\end{figure*}

As a multimodal dialog system, PerceptiveAgent is capable of audio modality perception and empathetic speech generation, which is achieved through the incorporation of prosodic information expressed in natural language. To capture prosodic features from speech inputs, we propose a novel speech caption model that aligns audio features with the latent space of a pre-trained language model. To enhance empathy and diversity of the simulated speech communication, a multi-speaker and multi-attribute vocoder is developed. This vocoder synthesizes speech by conditioning on both response contents and captions of speaking styles, resulting in more engaging and realistic dialogues.

\subsection{Speech Captioner} 
The speech caption model is designed to capture prosodic information and transcribe it as textual descriptions. It operates by encoding speech inputs by the speech encoder in ImageBind \citep{girdhar2023imagebind}, followed by description generation by the pre-trained GPT-2 decoder \citep{gpt2}. To bridge the gap between the speech encoder and the text decoder, we introduce a Querying Transformer (Q-former) pre-trained in BuboGPT \citep{zhao2023bubogpt}. This model is connected with a linear projection layer, which is subsequently followed by a text decoder. To effectively fine-tune this model, we integrate the following two fine-tuning strategies, while keeping the speech encoder frozen throughout the training procedure.

\subsubsection{Multi-modal Embedding Alignment} Prefix tuning is utilized to align the output of the Q-former with the latent space of the text decoder. A query vector with fixed dimensions is generated by the Q-former. These embeddings interact with each other through self-attention layers and with frozen audio features through cross-attention layers. To bridge the gap with the word embedding space, query embeddings are used as prefix vectors and attended to by the text decoder. This bottleneck architecture serves to compel the queries to extract the acoustic information that is most relevant to the textual descriptions.

\subsubsection{Instruction Tuning} To bridge the gap between the next-word prediction objective of the pre-trained decoder and the objective of acquiring multi-modal information conditioned on prefix sequences, instruction tuning is employed to train the speech captioner. We first construct 
an instructional dataset, where each instance comprises three elements: a query vector, an instruction, and a caption. The instruction is described as a natural language text sequence that specifies the task, serving to constrain the model’s outputs to align with desired response characteristics or domain knowledge. This provides a channel for humans to intervene with the model’s behaviors. Varied instructions are gathered using GPT-3.5-Turbo in this work. Additionally, the caption represents the desired output following the instruction, while the query vector is derived from acoustic representations. Throughout the training procedure, the parameters of the speech encoder are fixed, while the Q-former and text decoder remain trainable. During each inference process, instructions are randomly selected and incorporated into the generated sequence to enhance diversity and simulate human cognitive processes more effectively, thereby yielding more varied outputs.

\subsection{PerceptiveAgent}

Figure~\ref{fig:model} illustrates the overall framework of PerceptiveAgent,  a multi-modal dialogue system comprising three interconnected stages: Intention Discerning by the speech captioner, Comprehension through Sensory Integration by the LLM and Expressive Speech Synthesis by the MSMA-Synthesizer. PerceptiveAgent exhibits two key characteristics: (1) It leverages natural language to perceive and express acoustic information, and (2) It employs an LLM as the cognitive core in the system, to comprehend multi-modal contextual history and deliver audio responses.

\subsubsection{Caption for Intention Discerning} In the initial stage, a speech caption model is employed to interpret acoustic information from audio inputs. Each speech within the dialogue history is encoded into latent features by a frozen speech encoder. These features are then compressed into a query vector with fixed dimensions, sharing the same latent space as the word embedding of a text decoder. Conditioned on this query sequence and instruction prompt, a textual caption describing the speaking styles for each speech is deduced by the text decoder.

\subsubsection{Comprehension through Sensory Integration} Subsequently, an LLM module acting as the cognitive core is integrated into the system, where GPT-3.5-Turbo is employed. The transcribed textual content for each audio is merged with the previously generated caption before being fed into the LLM. Prompts in Appendix \ref{sec:appendix-1} and \ref{sec:appendix-2} are designed to effectively leverage the LLM's contextual understanding abilities. Upon recognizing speakers' intentions by assimilating both the contextual caption and content, the LLM deduces the relevant dialogue content and generates a caption describing how to articulate the derived content.

\subsubsection{Expressive Speech Synthesis} Finally, empathetic audio responses are  
synthesized by the  MSMA-Synthesizer, a Multi-Speaker and Multi-Attribute vocoder that is conditioned on the generated dialogue contents and captions. This vocoder is a modification of~\citep{nguyen2023expresso} 
to facilitate fine control over speech expressiveness. In addition to taking discrete speech units, speaker and style (emotion) as inputs, our vocoder introduces multiple prosodic attributes, including pitch, speed and energy. To synthesize each inference, the LLM's outputs of dialogue contents and captions are transformed into discrete units or attribute labels respectively, before being fed into the vocoder. Specifically, a text-to-unit (T2U) model is utilized to convert response contents into acoustic units with a Transformer machine translation structure \citep{vaswani2017attention}. Emotional and prosodic labels are recognized from response captions by sentence classifiers, accomplished with GPT-3.5-Turbo in this work, while the speaker label is randomly selected. 

The architecture of the vocoder comprises a speaker embedder, an attribute embedder and a HIFIGAN vocoder. The speaker embedder uses look-up tables to embed speaker identities, while a set of controllable attributes including speed, emotion, energy and pitch are embedded by the attribute embedder. To synthesize expressive speech, discrete units are initially embedded and up-sampled  through a series of blocks consisting of transposed convolution and a residual block with dilated layers. Prior to duration prediction, this up-sampled sequence is concatenated with the speed embedding. The speaker embedding and style embedding are subsequently concatenated to each frame in the up-sampled sequence, which is transformed to a mel-spectrogram by the HiFiGAN generator.

\section{Experiments}

\subsection{Experimental Setup}



\noindent\textbf{Datasets.} We train our speech captioner on the TextrolSpeech~\citep{ji2023textrolspeech} dataset, which consists of 236,220 pairs of captions and the corresponding speech samples. The captions in this dataset describe speaking styles in terms of five factors: gender, emotion, pitch, speed and energy. 

For the MSMA-Synthesizer, we reproduce a vocoder proposed in \citep{nguyen2023expresso} using the EXPRESSO, LJSpeech \citep{ljspeech17} and VCTK \citep{vctk} datasets. The EXPRESSO dataset is subsequently labeled by the speech captioner and GPT-3.5-Turbo to recognize attributes of pitch, speed and energy for each speech. We then utilize this labeled EXPRESSO dataset and the reproduced vocoder to train the MSMA-Synthesizer. We refer to the reading and conversation sections of EXPRESSO as Exp-R and Exp-I respectively. Additionally, a T2U model is trained on the same datasets with the MSMA-Synthesizer to maintain consistency in unit distribution.  

To evaluate the overall performance of our system, we utilize a speech dialogue dataset from MELD \citep{poria2018meld}. This dataset provides emotion labels for each sentence, which serve as ground truth labels for both response content and speech evaluation. The speeches in this dataset are recorded in realistic scenes with interruptions and environmental noise. In our evaluation, we only consider conversations with two speakers.

We utilize English datasets throughout the entire training process. As a consequence, PerceptiveAgent currently supports only the English language. However, it is noteworthy that  PerceptiveAgent can be readily expanded to accommodate multiple languages. Only the MSMA-Synthesizer module requires modification, as the language-agnostic nature of the speech captioner allows it to generate captions from various languages. Meanwhile, existing methods can recognize semantic contents and translate them into English. 

\noindent\textbf{Configurations.} We utilize the speech encoder in ImageBind \citep{girdhar2023imagebind}, the pre-trained Q-former in BuboGPT \citep{zhao2023bubogpt}, and the pre-trained GPT-2 \citep{gpt2}  to implement the speech captioner. 
Finetuning is conducted for 43,000 steps with a batch size of 16. For decoding, we use Top-k sampling with k=10 and set the minimum and maximum sequence lengths to 
20 and 50, respectively. We reproduce the vocoder for 400,000 steps with a batch size of 32 and learning rate of 0.0004, and train the MSMA-Synthesizer for 200,000 steps with a batch size of 32 and learning rate of 0.0004. The T2U model is structured as a sequence-to-sequence transformer with 4 encoder layers, 4 decoder layers, and 4 attention heads, with a dropout of 0.1. 
We utilize HuBERT~\citep{hsu2021hubert} with 2000 clusters to acquire units as targets\footnote{\url{{https://dl.fbaipublicfiles.com/hubert/hubert_base_ls960.pt}}}, provided by the textlesslib toolbox~\citep{kharitonov2022textlesslib}. Decoding is performed using Top-k sampling with k=10. All experiments are conducted on 4 NVIDIA GeForce RTX 4090 GPUs.

\subsection{Evaluation}

\noindent\textbf{Speech-GPT3.5.} We implement 
Speech-GPT3.5, a dialogue system focusing solely on linguistic information as a baseline. 
According to the textual history content recognized from the speech input, this system comprehends dialogue context with GPT-3.5-Turbo. After generating the response content, the audio response is synthesized by an off-the-shelf TTS (text-to-speech) model provided by OpenAI\footnote{\url{{https://platform.openai.com/docs/guides/text-to-speech}}}.

\noindent\textbf{Metrics.} The performance of PerceptiveAgent is evaluated in terms of two fundamental aspects: 1) \textit{cognitive empathy} demonstrates the ability to consider the perspective of speakers, reflected in the content of the response; and 2) \textit{affective empathy} exhibits the ability to emotionally understand and share the speaker's feelings, reflected in the prosody of the generated audio response. Cognitive and affective empathy 
are assessed by evaluating the quality of generated textual responses and audio responses, respectively. 

To evaluate the quality of dialogue text generation, we employ the BERTScore automatic evaluation metric proposed by \citet{zhang2019bertscore}, which computes a similarity score for each token in the candidate sentence with each token in the reference sentence. To evaluate the expressiveness of audio generation, we employ an expressive style classifier proposed by \citet{nguyen2023expresso} to recognize emotion labels for both generated and true speeches. 
Classification accuracy is used to measure the performance.
 
Besides, we evaluate the perception ability of the speech captioner on the validation and test datasets, which are split from the TextrolSpeech dataset. We approach this model as a multi-attribute classification task. Upon generating captions from speeches, the predicted labels for attributes including gender, emotion, pitch, speed and energy are determined by a sentence classifier, GPT-3.5-Turbo, while the true labels are provided in the TextrolSpeech dataset. Weighted metrics including precision, recall and 
F1-score are used to quantify the disparity between the predicted and true labels.

Moreover, the expressiveness of the speech synthesizer is assessed on the validation and test datasets split from the EXPRESSO dataset. We use the same 
expressive style classifier 
employed in affective empathy evaluation, to 
measure the preservation of emotion 
in the resynthesized speech. For evaluating the preservation of prosody, we compute the F0 Frame Error (FFE), which measures the percentage of frames with a deviation of more than 20$\%$ in pitch value between the input and resynthesized output.

\subsection{Result Analysis}

\subsubsection{PerceptiveAgent}


\begin{table}[t]
\centering
\begin{tabular}{rcc}
\toprule
                & BERTScore          & Accuracy       \\ \midrule
Speech-GPT3.5    & 53.03±10.20         & 0.74           \\
PerceptiveAgent & \textbf{54.36±9.25} & \textbf{21.89}   \\ 
-w/o captions & - & 16.53   \\ \bottomrule
\end{tabular}
\caption{Performance evaluation of PerceptiveAgent. 
BERTScore (\%) measures the quality of cognitive empathy in linguistic contents, while accuracy (\%) assesses the quality of affective empathy in acoustic responses. }
\label{tab:framework}
\end{table}

Table~\ref{tab:framework} presents the overall performance of PerceptiveAgent on cognitive empathy and affective empathy, evaluated on the generated content and audio, respectively. BERTScore measures the semantic similarity between the generated and real response contents, while accuracy assesses the similarity and diversity of emotions between the generated and real speeches. Overall, compared to Speech-GPT3.5, PerceptiveAgent demonstrates a strong ability in generating empathetic responses with a closer alignment to the dialogue context in terms of linguistic content and a higher expressiveness in acoustic information. Specifically, PerceptiveAgent achieves a slightly higher 
BERTScore 
than Speech-GPT3.5, primarily because our model can generate content that more accurately captures the speaker's intentions and contains more emotionally intense words. Additionally, PerceptiveAgent notably outperforms Speech-GPT3.5 in terms of accuracy, as 
the latter doesn't incorporate any emotion prompts during speech generation, thus maintaining a limited variety of prosody. Despite this, the accuracy of PerceptiveAgent still remains at a relatively moderate level. This is because the generated responses, while contextually appropriate, 
may not entirely align with the real responses in terms of semantics and emotions.

\subsubsection{Speech Captioner}


\begin{table*}
\centering
\begin{tabular}{c|cccccc}
\toprule
  \multirow{2}{*}{Attribute}                            & \multicolumn{3}{c}{Validation}                          & \multicolumn{3}{c}{Test}      \\ \cline{2-7} 
                    & Precision & Recall & \multicolumn{1}{c|}{F1-score} & Precision & Recall & F1-score \\ \hline
\multicolumn{1}{c|}{Gender}  & 99.3      & 97.5   & \multicolumn{1}{c|}{98.4}     & 99.3      & 98.6   & 99.0       \\
\multicolumn{1}{c|}{Emotion} & 85.8      & 85.4   & \multicolumn{1}{c|}{85.1}     & 87.3      & 87.1   & 86.8     \\
\multicolumn{1}{c|}{Pitch}   & 85.6      & 76.8   & \multicolumn{1}{c|}{80.4}     & 79.6      & 72.1   & 75.3     \\
\multicolumn{1}{c|}{Energy}  & 72.4      & 57.4   & \multicolumn{1}{c|}{63.1}     & 77.7      & 65.3   & 69.9     \\
\multicolumn{1}{c|}{Speed}   & 47.2      & 36.7   & \multicolumn{1}{c|}{41.3}     & 48.5      & 41.5   & 44.7     \\ 
\bottomrule
\end{tabular}
\caption{
Performance evaluation of the speech captioner. Precision, recall and F1-score (\%) are utilized to measure its generalization ability on both the validation and test sets. Predicted labels are obtained through semantic classification on the generated captions, while the true labels are derived from the TextroSpeech dataset.}
\label{tab:caption}
\end{table*}

Table~\ref{tab:caption} evaluates the speech captioner's generalization performance on both the validation and test sets. Overall, it is evident that that the model achieves the highest F1-score for gender, followed by pitch and emotion. This underscores the model's proficiency in accurately discerning 
these attributes from input speech. Besides, both gender and emotion exhibit closely aligned precision and recall metrics, affirming the model's predictive prowess for these attributes. 
Meanwhile, there exists a notable disparity between precision and recall 
when predicting energy, indicating variable performance and a tendency towards confident predictions. 
Conversely, the model's performance in predicting speed is unsatisfactory, which can be attributed to the imbalanced distribution of speed in the training dataset, with over 60\% of samples labeled as ``neutral''.

We also discuss how errors in speech processing are affected by demographics of the speakers. Table ~\ref{tab:comparison} compares the performance of the speech captioner across genders, which represents the most prevalent factor. The F1-score on male speech surpasses that on female speech in terms of pitch, energy and speed, despite the comparable sample sizes for male and female groups (8634 VS. 8983). This demonstrates a variation in the model's performance depending on the gender of the speakers. 

\begin{table*}
\centering
\begin{tabular}{c|cccccc}
\toprule
  \multirow{2}{*}{Attribute}                            & \multicolumn{3}{c}{Male}                          & \multicolumn{3}{c}{Female}      \\ \cline{2-7} 
                    & Precision & Recall & \multicolumn{1}{c|}{F1-score} & Precision & Recall & F1-score \\ \hline
\multicolumn{1}{c|}{Emotion}  & 84.3      & 85.4   & \multicolumn{1}{c|}{84.2}     & 87.4      & 85.5   & 86.0       \\
\multicolumn{1}{c|}{Pitch} & 88.2      & 82.8   & \multicolumn{1}{c|}{85.3}     & 84.8      & 71.0   & 75.9     \\
\multicolumn{1}{c|}{Energy}   & 74.4      & 60.0   & \multicolumn{1}{c|}{65.0}     & 71.2     & 54.9   & 60.9     \\
\multicolumn{1}{c|}{Speed}  & 46.4     & 43.1   & \multicolumn{1}{c|}{44.6}     & 48.0      & 30.6   & 37.3     \\
\bottomrule
\end{tabular}
\caption{Comparison of the speech captioner's performance across genders.}
\label{tab:comparison}
\end{table*}

\subsubsection{MSMA-Synthesizer}


\begin{table}
\centering
\begin{tabular}{c|cc|c}
\toprule
 \multirow{2}{*}{Method}                & \multicolumn{2}{c|}{Accuracy} & FFE       \\
 & Exp-R         & Exp-I         & Exp       \\ \hline
GT     & 91.9        & 75.1       & -         \\
EXPRESSO         & \textbf{87.9}       & 67.0       & \textbf{0.17±0.12} \\
MSMA & 83.8       & \textbf{70.8}       & 0.39±0.16 \\ 
\bottomrule
\end{tabular}
\caption{Preservation evaluation of MSMA-Synthesizer. 
Accuracy (\%) is evaluated on EXPRESSO read (Exp-R) and conversation (Exp-I) dataset. F0 Frame Error (FFE) is calculated on EXPRESSO (Exp). GT represents the results of automatic metrics calculated on real audio. EXPRESSO and MSMA refer to the synthesizers in EXPRESSO and PerceptiveAgent respectively.}
\label{tab:synthesizer}
\end{table}

Table~\ref{tab:synthesizer} assesses the MSMA-Synthesizer's ability to preserve emotion and prosody features on the test set, where the EXPRESSO Synthesizer is reproduced by 
us. The ``GT'' method represents the results of automatic metrics calculated on real audio. Clearly, the MSMA-Synthesizer achieves higher accuracy on the read dataset compared to EXPRESSO. This suggests that 
an integration of multiple attributes into speech synthesis can more effectively enable the model to synthesize emotionally expressive audio in dialogue scenarios, meeting the requirements of our system. However, there is a decrease in accuracy on the Exp-R dataset, which is relevant to the less apparent variation in prosody with emotional transitions. Additionally, 
in terms of FFE, it can be observed that incorporating multiple attributes into the MSMA-Synthesizer may lead to some degradation in speech synthesis quality. However this degradation 
remains within an acceptable range.

\subsubsection{Ablation Study}
\noindent\textbf{Effectiveness of Captions.}
The last line in Table~\ref{tab:framework} demonstrates the effectiveness of captions in PerceptiveAgent. The system without captions synthesizes speech using randomly selected labels for all four speaking attributes (pitch, speed, energy, and emotion), while maintaining the same response contents as the PerceptiveAgent. It is evident that the PerceptiveAgent outperforms the system without captions, highlighting the effectiveness of captions in generating speech with affective empathy.

\noindent\textbf{Effectiveness of Style Factors.}
To discern the discrete impact of distinct speaking style factors, we conduct an ablation experiment by systematically varying each factor while maintaining the others at their default values. Table~\ref{tab:ablation} presents that the model with style remained achieves the highest accuracy and the lowest FFE, while the models with the other factors exhibit similar performance. This underscores the predominant contribution of style to the effectiveness of expressive speech synthesis.

\begin{table}
\centering
\begin{tabular}{r|cc|c}
\toprule
 \multirow{2}{*}{Method}                & \multicolumn{2}{c|}{Accuracy} & FFE       \\
 & Exp-R         & Exp-I         & Exp       \\ \hline
GT     & 91.9        & 75.1       & -         \\
EXPRESSO         & \textbf{87.9}       & 67.0       & \textbf{0.17±0.12} \\
MSMA & 83.8       & \textbf{70.8}       & 0.39±0.16 \\ 
-style & 82.2       & 69.0       & 0.40±0.16 \\ 
-speed & 31.8       & 9.2       & 0.44±0.13 \\
-energy & 31.0       & 9.1       & 0.44±0.13 \\
-gender & 30.8       & 8.7       & 0.44±0.13 \\
-pitch & 30.7       & 7.4       & 0.43±0.13 \\ 
\bottomrule
\end{tabular}
\caption{Performance of the MSMA-Synthesizer conditioned on single speaking style factors.}
\label{tab:ablation}
\end{table}

\section{Case Study}

\begin{figure*}[h]
    \centering
    \subfigure[Contradictory Example]{
    \label{time.1}
    \includegraphics[width=11cm]{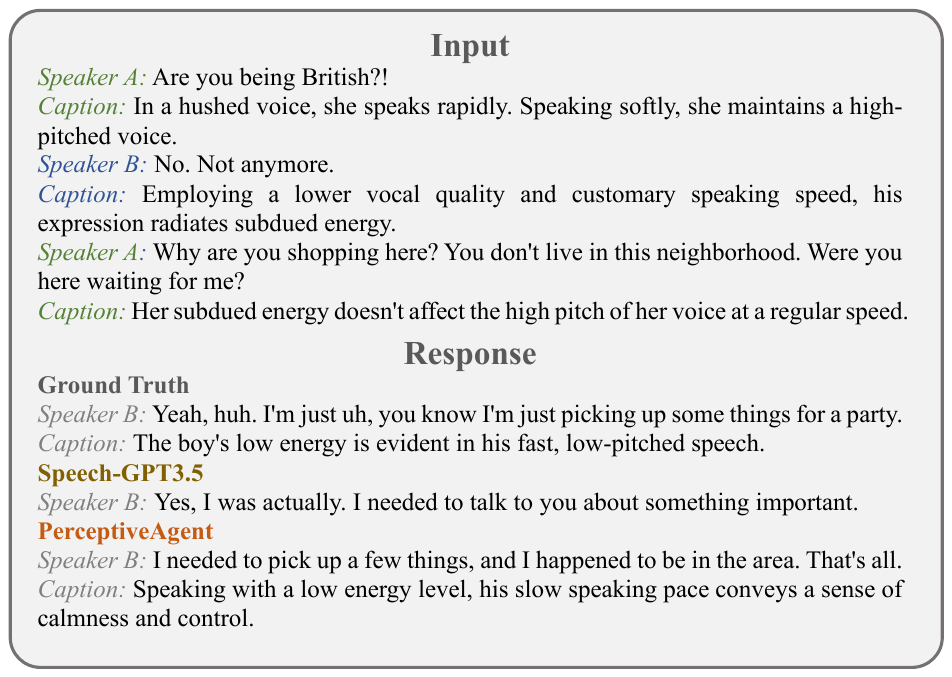}}
    \quad
    \subfigure[Consistent Example]{
    \label{time.2}
    \includegraphics[width=11cm]{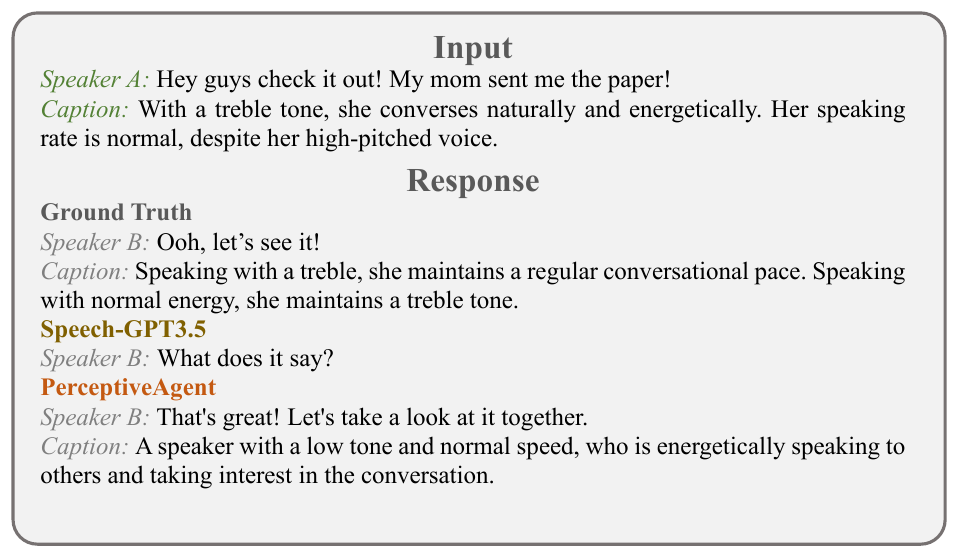}}
    \caption{Cases comparing the response quality between Speech-GPT3.5 and PerceptiveAgent. }
    \label{fig:cases}
\end{figure*}

Figure~\ref{fig:cases} presents two cases comparing the response quality between Speech-GPT3.5 and PerceptiveAgent. It demonstrates that by explicitly incorporating acoustic information through captions, the LLM can more accurately comprehend the speaker's intentions and generate more accurate and contextually appropriate responses. The first and second examples illustrate scenarios where the speaker's intention either contradicts or aligns with the linguistic contents, respectively.

The first example in Figure \ref{fig:cases} (a)  depicts an unplanned meeting conversation between two friends.  Analyzing solely from the textual contents, 
it is suggested that the speaker B is extremely excited and delighted about this conversation. However, a closer examination of the key words of ``lower vocal'' and ``subbed energy'' in speaker B's caption reveals an evasive attitude towards the situation. Consequently, when confronted with speaker A's question, ``Were you here waiting for me?'', it can be inferred that speaker B is not inclined to engage in extensive conversation. The absence of nuanced captions poses a challenge for Speech-GPT3.5, leading to a misinterpretation and generating a response that implies a strong desire to continue the conversation. In contrast, PerceptiveAgent provides a response in accordance 
with the underlying meaning. Therefore, despite potential inconsistencies between linguistic contents and speaker intentions 
disrupting the accuracy of dialogue context understanding, PerceptiveAgent, with the aid of captions, can effectively capture the speaker's intent by correctly discerning the acoustic information of speech.

In the second example in Figure \ref{fig:cases} (b), the speaker A receives a paper from his mother and intends to share it with his friends. It can be inferred that he is highly excited at the moment, as evidenced by the key words ``treble tone'' and ``energetically'' in the caption. Recognizing speaker A's excited mood, the response generated by PerceptiveAgent mirrors the same enthusiasm and curiosity, aligning well with the ground truth. However, Speech-GPT3.5 fails to perceive speaker A's excitement and merely raises the question in a bland manner. Thus, in scenarios where the textual contents coincides with the speaker's intent, our model can also provide responses that correspond to the context of the conversation.

\section{Conclusion}

In this paper, we propose PerceptiveAgent, an empathetic multi-modal dialogue system capable of accurately discerning the speaker's intentions through the integration of perceptive speech captions and to respond with nuanced and expressive spoken dialogues. Specifically, 
PerceptiveAgent comprises three cascaded modules: a speech captioner for intention discernment, an LLM for comprehension through sensory integration, and an MSMA-Synthesizer for expressive speech synthesis. Initially, the system employs a perceptive captioner model to capture acoustic features from each speech within dialogues. Subsequently, an LLM module serves as the cognitive core, generating relevant response content with a caption conditioned on the comprehension of the speaker's intentions. An MSMA-Synthesizer is then developed to synthesize expressive speech. Experimental results indicate PerceptiveAgent's 
strong ability in empathetic response generation, closely aligning with the dialogue context in terms of linguistic contents and exhibiting high expressiveness in acoustic information. Additionally, a case study demonstrates PerceptiveAgent's capability to accurately identify the speaker's intentions in scenarios where the literal interpretations of words are either contrary to or inconsistent with the speaker's true feelings.

\section{Limitations}

Although PerceptiveAgent excels at providing empathetic responses in terms of both linguistic and acoustic contents, several limitations can be observed in this system: 1) \textbf{Dataset Limitation}: PerceptiveAgent's perception ability is currently constrained by the comprehensiveness of the training dataset in describing speech information. Presently, it is unable to discern speaker identity and background noise from speech; 2) \textbf{Time Delay Limitation}: PerceptiveAgent is a system cascaded by three interconnected components, which introduces accumulated delays to the response time, 
and 3) \textbf{Length Limitation}: The maximum token length in LLMs may limit the multi-turn dialogue.

\section*{Acknowledgements}
This research was supported by the National Natural Science Foundation of China (Grant No. 62276245).



\onecolumn
\appendix

\section{Prompt for Dialogue Generation with Captions}
\label{sec:appendix-1}

\begin{figure*}[htbp]
    \centering
    \includegraphics[width=0.9\textwidth]{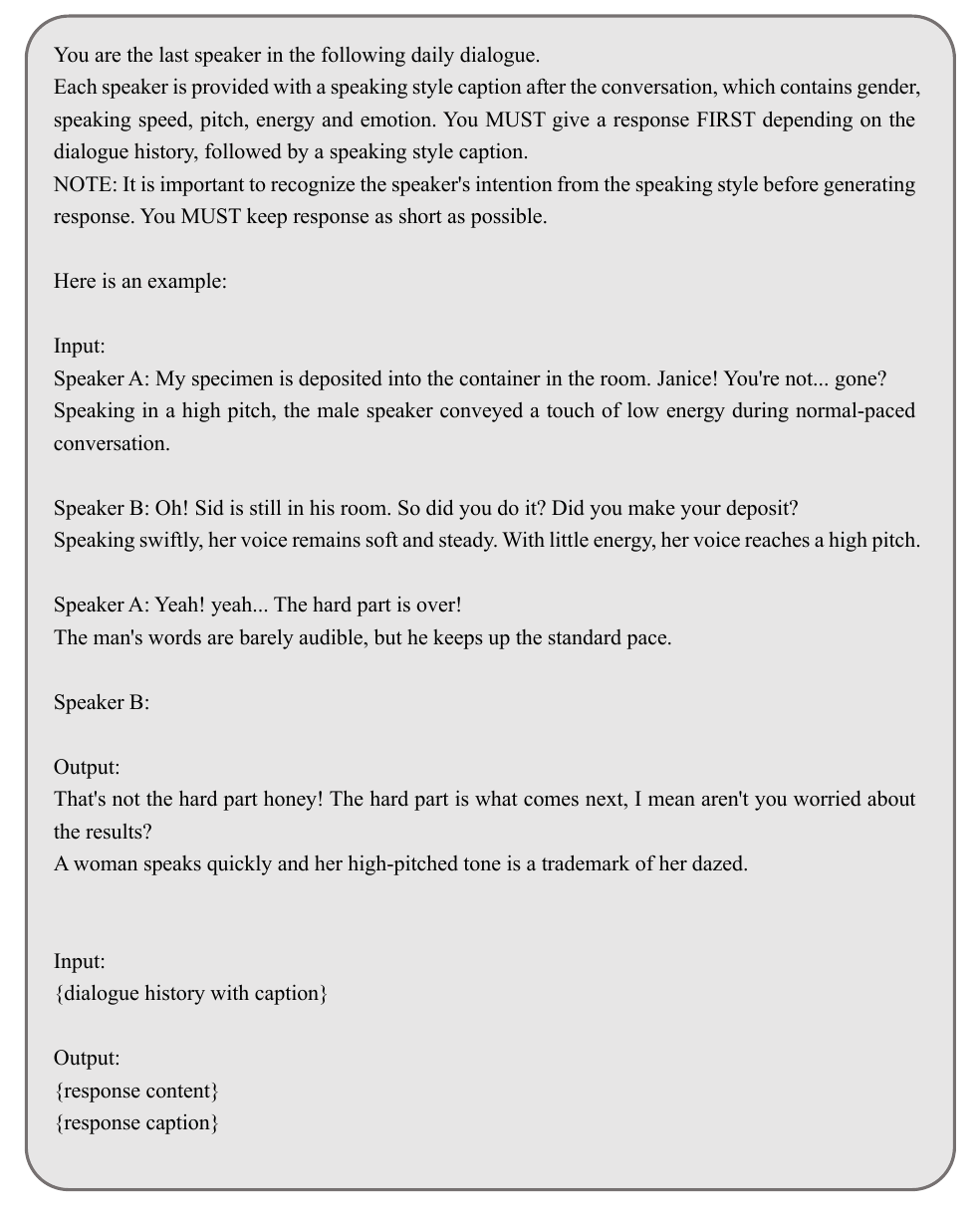}
    \label{fig:prompt_caption}
\end{figure*}

\clearpage

\section{Prompt for Dialogue Generation without Captions}
\label{sec:appendix-2}

\begin{figure*}[htbp]
    \centering
    \includegraphics[width=0.9\textwidth]{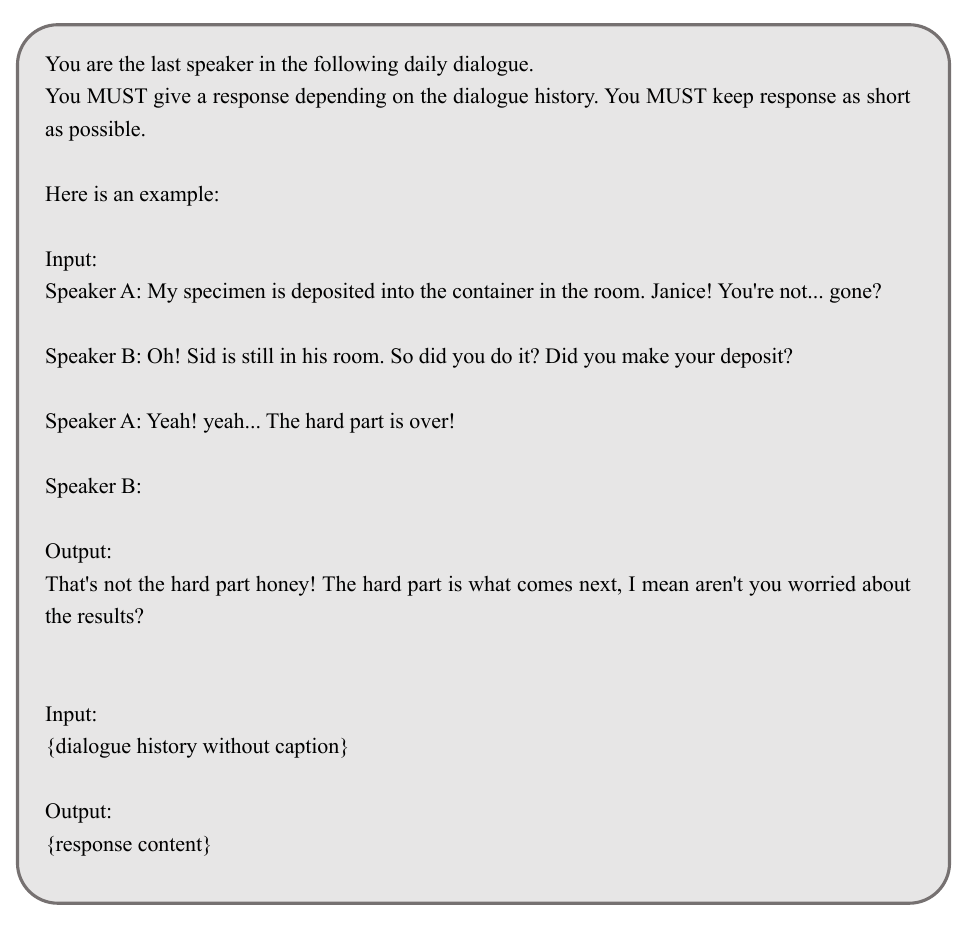}
    \label{fig:prompt_text}
\end{figure*}

\twocolumn
\end{document}